\documentclass{article}

\usepackage[final]{nips_2017}

\usepackage[utf8]{inputenc} 
\usepackage[T1]{fontenc}    
\usepackage{hyperref}       
\usepackage{url}            
\usepackage{booktabs}       
\usepackage{amsfonts}       
\usepackage{nicefrac}       
\usepackage{microtype}      
\usepackage[pdftex]{graphicx} 
\usepackage{placeins}
\usepackage{booktabs}

\setcitestyle{numbers}

\title{Pre-training Attention Mechanisms}

\author{
  Jack Lindsey\thanks{Code available on GitHub at github.com/jlindsey15/DRAM} \\
  Stanford University\\
  \texttt{jacklindsey@stanford.edu} \\
}

\begin{document}

\maketitle

\begin{abstract}
  Recurrent neural networks with differentiable attention mechanisms
  have had success in generative and classification tasks.  We show that 
  the classification performance of such models 
  can be enhanced by guiding a randomly initialized model to attend to salient
  regions of the input in early training iterations.  We further show that, if explicit heuristics for
  guidance are unavailable, a model that is pre-trained on an unsupervised reconstruction task
  can discover good attention policies without supervision.  We demonstrate that increased efficiency 
  of the attention mechanism itself contributes to these performance improvements.  Based on these insights, we 
  introduce bootstrapped glimpse mimicking, a simple, theoretically task-general method of more effectively training attention models.  Our work draws inspiration from and parallels results on human learning of attention.
  
\end{abstract}

\section{Introduction}

Humans process visual data by selectively attending to different areas, where attention may be construed as the differential processing of simultaneous
sources of information \citep{johnston1986selective}.   Previous studies of attention in recurrent neural networks successfully model some attributes of this sequential, glimpse-based approach \citep{gregor2015draw,mnih2014recurrent,ranzato2014learning,alexe2012searching,butko2009optimal,denil2012learning,larochelle2010learning,paletta2005q,tang2014learning,zheng2015neural,ba2014multiple,sermanet2014attention}.  However, these models
have been trained from scratch, in the sense that the network must simultaneously
learn how to choose areas of focus and how to process them once chosen.
A human attempting a classification task, on the other hand, benefits from
substantial prior knowledge of how to select salient regions of an input to attend
to \citep{kim2011prior,shiffrin1977controlled,schneider1977controlled}.

It seems reasonable that imbuing a model with such prior knowledge would allow it to learn to complete the relevant task more quickly.  Indeed, unsupervised pre-training is known to assist in the training of deep neural networks \citep{erhan2010does,erhan2009difficulty}.  However, a priori, one might not expect a "pre-trained" model to outperform one trained from scratch, if both were given unlimited training time.  Nevertheless, there are reasons to suspect such a phenomenon could indeed occur.  We were motivated principally by the insight that recurrent models of attention influence their own training inputs to a high degree.  More precisely, the glimpses used to train the model are dependent on the model parameters at a given time step.   Thus, a model with a policy that attends primarily to salient patches of input might learn more efficiently from training data, and avoid noisy training signals that have been shown to impair a wide variety of models \citep{kalapanidas2003machine, chen2014big}.  

In this work, we model prior knowledge of attention policies -- both task-general and task-specific -- in several ways and find them to
be effective in improving classification accuracy after training converges.  
We begin with examples that illustrate the effect clearly on a limited class of problems, and transition towards
more robust methods.  A promising approach
for practical use,"bootstrapped glimpse mimicking," appears near the end of this paper.  

Our methods are inspired by the relevant cognitive science literature, and our results map well to psychological findings. In particular, we highlight the effects of prior task-specific domain knowledge on attention \citep{kim2011prior, lin2014domain}, task-general attention pre-training \citep{huang2012task, tang2009attention, jiang2015task}, and attention policy transfer and task-specific adaptation \citep{schroth2000effects,cantor1965transfer, lin2014domain, van1997task}.  Furthermore our bootstrapping algorithm finds parallels in models of other cognitive phenomena \citep{brady2003bootstrapped, goodman2011learning, gentner2010bootstrapping, piantadosi2012bootstrapping}.

\section{Model}

We used much of the structure of the DRAW model described in \citet{gregor2015draw}.  Our principal deviations from that model were as follows.
First, we did not use attention while "writing" reconstructions.  In fact, our network did not build up reconstructions
up over time; rather, the reconstruction output was entirely dependent on the output of the decoder network at
the final time step.  This modification was necessary to 
enable pre-training by allowing the classifier to build off the internal representations learned by the reconstructor. Second, our model was not variational; that is, no noise was added to the encoding.  We did this to simplify the model, as our intent was not to reach maximum performance on generative tasks but rather to explore methods of imparting prior knowledge on an attention mechanism.

\subsection{Attention Mechanism}

To enable faster training times, we chose to use a fully differentiable attention mechanism.
Thus the model may be trained end-to-end using backpropagation, rather than more computationally
expensive reinforcement learning.  The mechanism used is the same
as the one developed for the DRAW model.  It consists of a square array of regularly spaced Gaussian
filters applied to
the input.  Five parameters define the area of focus at a given time step.  Two specify the center of the
array, one determines the variance of the Gaussian filters, one determines the spacing of the filters, and
a fifth parameter multiplies the reading intensity by a constant.  More details are available in \citet{gregor2015draw}

\subsection{Network Architecture}

\begin{figure}[h]
  \centering
  \includegraphics[width=0.4\linewidth]{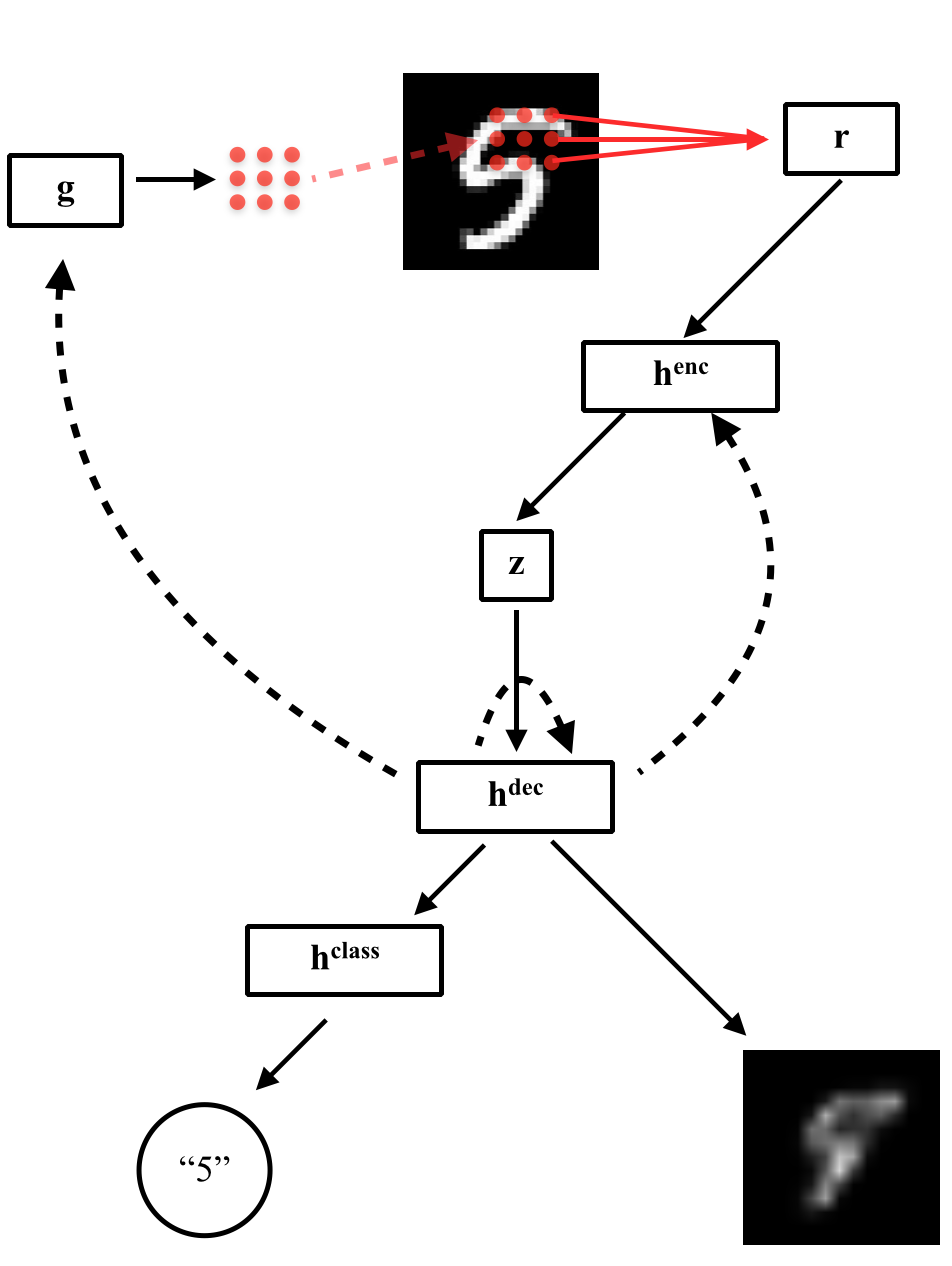}
  \caption{Schematic of model architecture.}
\end{figure}

The network consists of an encoder and a decoder, both recurrent LSTM cells.  The network runs for
a fixed number of time steps, T (we set T = 10 in all experiments).  At each step, the five glimpse parameters
are obtained by applying a linear transformation to the output $h^{dec}_{t-1}$ of the decoder cell (whose contents are
zero initially).  The attention mechanism described above then produces a "glimpse," $r_t$, based on these parameters. 
The encoder cell receives $r_t$ and $h^{dec}_t$ as inputs, and outputs $h^{enc}_t$, which is fed through another linear layer
to produce the encoding, $z_t$. 

The decoder takes $z_t$and $h^{dec}_{t-1}$ as inputs and outputs $h^{dec}_t$.  For the classification task, after T time steps,
$h^{dec}_T$ is fed through a linear layer with a ReLU nonlinearity to yield $h^{class}$, which is in turn fed through a linear
layer followed by a softmax function to output classification probabilities.  For the reconstruction task, $h^{dec}_T$ is input to a linear
layer followed by a sigmoid (since MNIST pixel values are represented as scalars between 0 and 1).  

Training occurred in batches of size 100. All weights were initialized with the Xavier uniform random initializer \citep{glorot2010understanding}.

\begin{table}[t]
  \caption{Network Hyperparameters}
  \label{sample-table}
  \centering
  \begin{tabular}{llllll}
    \toprule
    \cmidrule{1-2}
    Task     & T     & LSTM cell size & z size & $h^{class}$ size & read size  \\
    \midrule
    100 x 100 Translated MNIST & 10 & 256 & 10 & 256 & 5 x 5    \\
    Regular MNIST     & 10 & 256 & 10 & 256 & 2 x 2      \\
    \bottomrule
  \end{tabular}
\end{table}

\section{Explicitly Guided Attention}

Motivated by relevant work in the psychology literature \citep{kim2011prior, lin2014domain}, we consider the effect of explicitly directing where the learner should
focus its attention based on prior domain knowledge. 
We consider 100 x 100 Translated MNIST task, in which the digit to be classified occupies only a small region of the image input, allowing us to test a simple domain-specific heuristic for incentivizing efficient attention policies.

\subsection{Methods}

We proceeded by pre-training a model using a teaching signal
that rewards glimpses centered near the digit's location. We used, as the loss
signal L, the total (across all time steps) squared distance between the position $x_{dig}$ of the center of the MNIST digit in the image , and the position $x_{att}$ of the center of the attention grid.  Additionally, we applied L1 regularization, so that the error
signal was computed as:
\[L = ||x_{dig} - x_{att}||_{l2} + 0.001 \cdot ||h_{dec}||_{l1}\]

where subscripts refer to the l2 and l1 norms.

Pre-training proceeded for 2000 training batches, after which we trained the pre-trained model directly on the MNIST classification task. 

\subsection{Results}

All performance figures (Table~\ref{performance}) were obtained using a separate test dataset containing 10000 examples.  Test Error was computed after every 1000 training
batches.  The values listed reflect an average over ten computations of testing error after accuracy stopped improving.  We observe that guided models reached
lower minimum error than unguided models (1.4\% vs. 1.9\% ).  They also converged to high performance much more quickly
(3000 training batches until 95\% accuracy vs. 42000).  No additional performance benefits were obtained by increasing the
number of pre-training batches.

\section{Reconstruction Pre-training}

Of course, such an explicit heuristic for guiding glimpse policy is often not available.  For example, in the traditional MNIST
task where the digits occupy most of the input, it is difficult to say a priori where we would like the model to attend to. We seek a way of modeling the task-general training of attention mechanisms.  Such phenomena are known to be significant in human attentional behavior \citep{huang2012task, tang2009attention, jiang2015task}, and are distinct from more ``effortful'' task-specific attention learning \citep{bruya2010effortless}. For simple inputs like MNIST or Translated MNIST, we conjectured that training the model
to reconstruct its input would approximate this effect, forcing the model to learn how to attend to generally salient 
features of the input before it is trained on a classification task.

\begin{figure}[h]
  \centering
  \includegraphics[width=1\linewidth]{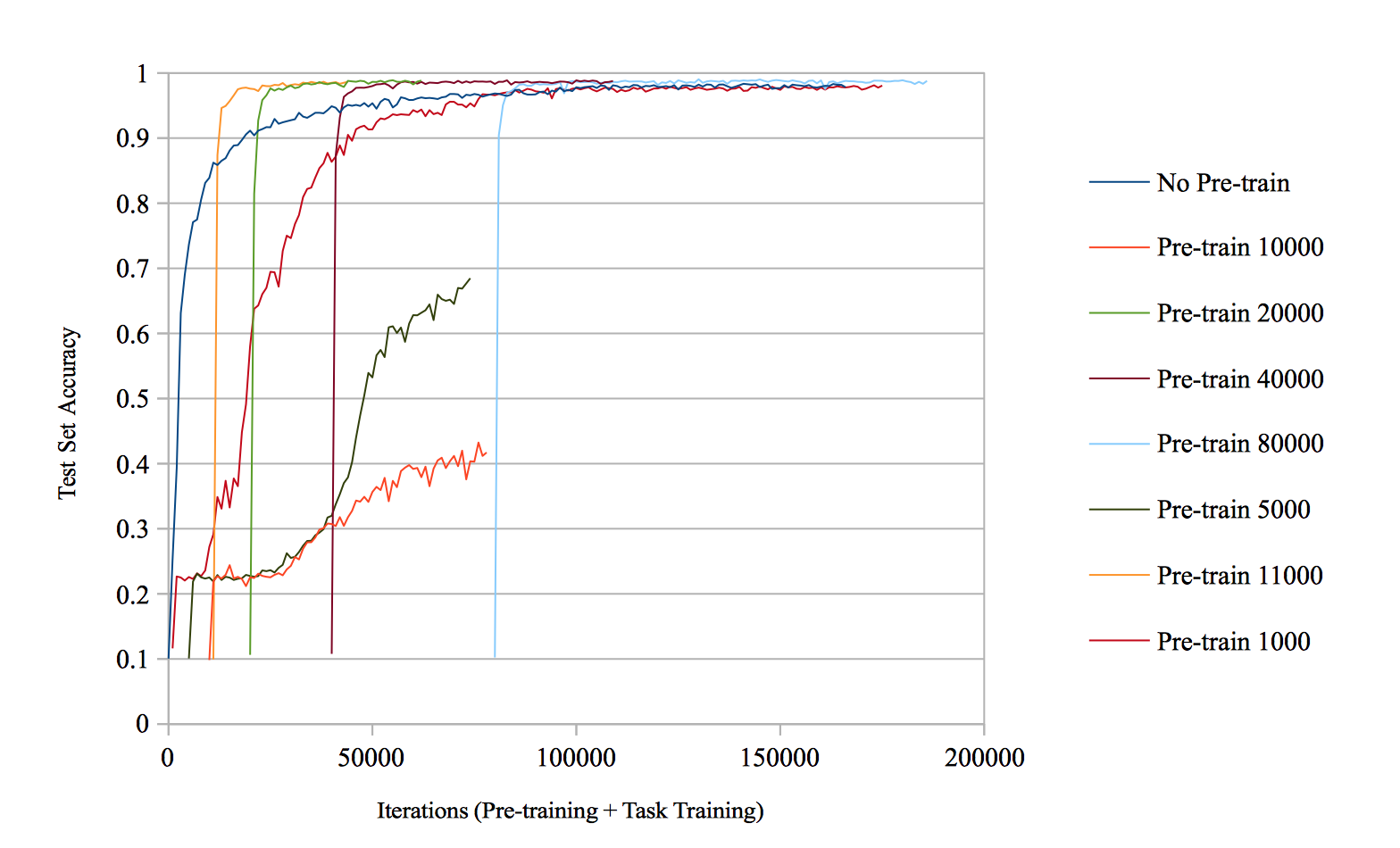}

  \caption{Comparison of learning trajectories after employing reconstruction pre-training for different numbers of iterations.  The starting point for each trajectory corresponds the the number of pre-training iterations in that trial.  Note that the trials cluster into (i) runs with little or no pre-training, which give good performance, (ii) runs with sufficient (at least 11,000 iterations) pre-training, which give better performance, and (iii) runs with insufficient pre-training, which give poor performance. Best viewed in color.}
  \label{performancegraph}
\end{figure}

 We used a binary cross-entropy loss function, applied to the training input and the network's attempted reconstruction,
as our error signal.  We allotted 20000 pre-training epochs  for the translated MNIST task and 15000 for regular MNIST.  As before,  sufficient pre-training yielded improvement in both asymptotic performance and training time, but no
additional performance improvements were obtained by increasing the number of pre-training batches.  Insufficient pre-training actually reduced performance in some cases.  \footnote{Note that models
trained on regular MNIST were given a smaller read window, and thus they should not necessarily be expected to perform better
than on translated MNIST.} \\
Results for each type of pre-training are given in Table~\ref{performance}, and learning trajectories in Figure~\ref{performancegraph}. Reconstruction pre-training results were consistent for 20,000 iterations and above.  Each experiment was repeated five times, and results are given to tenths digit precision, where they were consistent across trials.

\begin{table}[t]
  \caption{Performance Comparison (Direct Pre-training)}
  \label{performance}
  \centering
  \begin{tabular}{llllll}
    \toprule
    \cmidrule{1-2}
    Training Method & Error  \\
    \midrule
    \textbf{100 x 100 Translated MNIST} \\
    No Guidance & 1.9\%    \\
    Heuristic-based Pre-training & 1.4\% \\
    Reconstruction Pre-training & 1.4\%     \\
    \textbf{Regular MNIST} \\
    No Guidance & 2.6\%    \\
    Reconstruction Pre-training & 2.2\%     \\  
    
    \bottomrule
  \end{tabular}

\end{table}

\section{Measuring Glimpse Efficiency}

It is worth investigating what drives the performance improvements yielded by guided attention.  In particular, we consider the \emph{efficiency} of a learned attention policy, a concept explored in the human setting by, \citet{van2002redirecting} and \citet{castiello1990size}, among others. We must distinguish between performance improvements due to increased efficiency of the attention mechanism itself as opposed to improved ability to process information after selecting a focus region.  In this section, we demonstrate that at least a significant
portion of the accuracy gains obtained from guided or pre-trained attention can be attributed to more information-efficient selection of focus regions.

\begin{table}[t]
  \caption{Glimpse Efficiency}
  \label{glimpseefficiency}
  \centering
  \begin{tabular}{llllll}
    \toprule
    \cmidrule{1-2}
    Training Method / Duration  & Glimpse Quality  \\
    \midrule
    \textbf{100 x 100 Translated MNIST} \\
    Reconstruction, 80000 batches & 93.9\%    \\
    Classification, 80000 batches & 95.4 \% \\
    Reconstruction 40000 batches + classification 40000 batches & 98.6\%     \\
    \textbf{Regular MNIST} \\
    Reconstruction, 25000 batches & 96.8\% \\
    Classification, 25000 batches & 96.0\% \\
    Reconstruction 15000 batches + classification 10000 batches & 97.4\% \\
    \bottomrule
  \end{tabular}
\end{table}

\begin{figure}[h]
  \centering
  \includegraphics[width=0.4\linewidth]{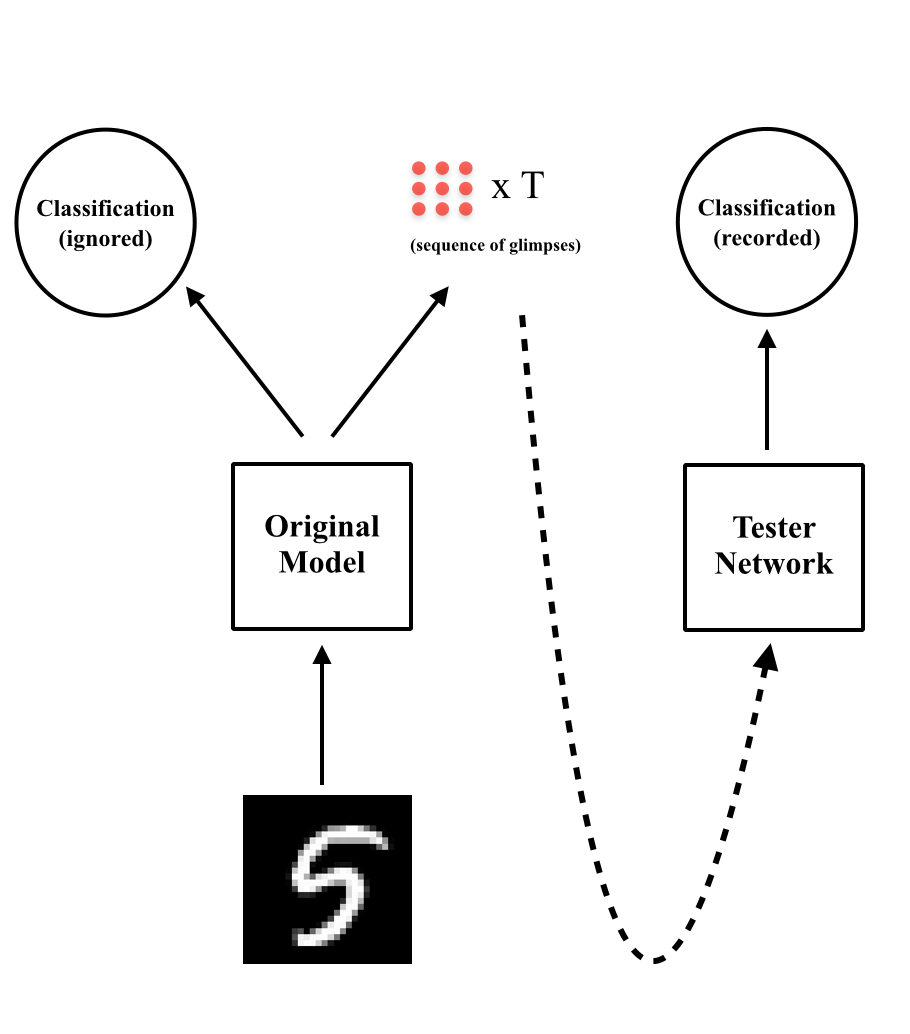}
  \caption{High-level procedure for measuring glimpse efficiency.}
\end{figure}

In our model, the classification prediction and the glimpse selection are functions of the same hidden representation,
making it difficult to separately analyze the quality of its glimpse selection and the quality of its image processing.
We analyze focus window selection efficiency by the use of a read-out function, as in \citet{hong2016explicit} and \citet{zamir2016generic}, designed to measure the information about the task categories that can be extracted from the set glimpses taken.  We fed images through
a trained model and stored, at each time step, the output $r_t$ of the glimpse operation, along with the parameter vector
$g_t$ used in the glimpse operation.  These were fed as training inputs to a "tester" network, a recurrent LSTM network with no attention mechanism.  The ability of the tester network to achieve high classification accuracy given the glimpses selected by the original
model provides a measure of the efficiency of the model's glimpses.  Note that these figures (Table~\ref{glimpseefficiency}) are useful only relative to one another.  \footnote{They
are not suitable for comparison with other values in this paper, as these are obtained using a different network architecture.}

As measured by the tester network, models pre-trained on reconstruction and then trained
on classification had developed more efficient glimpse policies than models trained on either task alone (even controlling for
total number of training batches).  Thus, we may conclude that improved selection of glimpses does indeed fuel the improved
performance of our guided models.  Furthermore, we rule out the hypothesis that reconstruction pre-training itself somehow
discovers these efficient glimpse policies.  Rather, the combination of prior knowledge gained through reconstruction
pre-training and the ensuing task-specific training is necessary to see performance benefits.  Hence, task-specific training is still essential to developing high-performing attention policies.  This result is simultaneously consistent with psychological findings on the benefits of attentional pre-training \citep{schroth2000effects,cantor1965transfer} and on the importance of role of domain-specific attention policy adaptation \citep{lin2014domain, van1997task}.

\section{Bootstrapped Glimpse Mimicking}

Though reconstruction pre-training can succeed in improving performance, it has a few undesirable attributes.  First, as already mentioned, pixel reconstruction may fail to be of use for complex
data.  Adversarial reconstruction has the potential to alleviate this problem, but at great computational expense.  Second, using a pre-trained model might bias the representations learned by the network when it is trained on classification.  Here, we investigate a simple method that still takes advantage of prior knowledge in the
attention mechanism but allows the model of interest to be trained from scratch.  The new model is biased to learn glimpse patterns similar to those of the pre-trained model ("mimicking"), but not necessarily similar representations.  Our approach also
allows the network architecture used to obtain prior knowledge about efficient glimpse policies to be separate from the network
architecture used in the classification task.  One can imagine scenarios in which different architectures are better suited to these
different problems.

\subsection{Glimpse Mimicking}

To illustrate glimpse mimicking, we proceed as follows.  First, we trained a model on the reconstruction task, as before.  Then a new model with randomly initialized weights was trained
on the MNIST classification task.  However, it also received a supplementary, decaying error signal penalized any difference between
the vector of glimpse parameters chosen by the model at a time step, $g^1_t$, and those that would have been chosen by the pre-trained
model, $g^0_t$.  We use L to denote errors.

\[L_{tot} = (L_{class}) + \alpha^{\lambda \cdot i} \cdot \frac{1}{T} \sum_t ||g^1_t - g^0_t||_{l2}\]

with $\lambda = 0.01$, and i representing the number of the training batch.

We refer to this technique as "glimpse mimicking" and call the pre-trained model the "teacher network."  In our experiments, glimpse mimicking using a model pre-trained on reconstruction could yield comparable performance benefits to direct pre-training with $\alpha = 0.95$.

\subsection{Bootstrapping}

Given our discovery that reconstruction pre-training itself does not discover efficient glimpse policies until classification training
is added on, we might wonder whether the reconstruction step is necessary at all.  Indeed, we may train a model from scratch
\emph{on classification} and use it as the teacher network for glimpse mimicking.  In this manner we may bootstrap the process of learning, a phenomenon already thought to play a role in human learning of object detection \citep{brady2003bootstrapped}, causality \citep{goodman2011learning}, analogy \citep{gentner2010bootstrapping}, and number sense \citep{piantadosi2012bootstrapping}. We trained the teacher network for 5000 epochs on Regular MNIST and 80000 on Translated MNIST.  Our empirical results
suggest that this method, albeit sensitive to the $\alpha$ hyperparameter defined above, is effective.  Furthermore, it makes no
assumptions about any characteristics of the input data or even the nature of the task being completed.

\begin{table}[t]
  \caption{Glimpse Mimicking Performance}
  \label{sample-table}
  \centering
  \begin{tabular}{llllll}
    \toprule
    \cmidrule{1-2}
    Method  & Error  \\
    \midrule
    \textbf{100 x 100 Translated MNIST} \\
    No Guidance & 1.9\% \\
    Reconstruction pre-train mimicking, $\alpha = 0.95$ & 1.5\% \\
    Bootstrapped mimicking, $\alpha = 0.95$ & 1.5\% \\
    Bootstrapped mimicking, $\alpha = 0.50$ & 2.7\% \\
    \textbf{Regular MNIST} \\
    No Guidance & 2.2\% \\
    Reconstruction pre-train mimicking, $\alpha = 0.95$ & 1.9\% \\
    Bootstrapped mimicking, $\alpha = 0.95$ & 2.3\% \\
    Bootstrapped mimicking, $\alpha = 0.50$ & 1.9\% \\
    \bottomrule
  \end{tabular}
\end{table}

\section{Discussion}

We have made a few noteworthy discoveries.  The first is that recurrent models of attention can often, when trained from scratch on
an image classification task, fail to perform optimally.  Though we have not established whether these suboptimal solutions represent
local minima in parameter space or simply a region of insufficiently large gradients, we know they are suboptimal as we have
found better-performing policies.  These superior policies, in our experiments, were only reachable through some form of pre-training.  We explored several methods of pre-training, including the use of highly supervised, explicitly defined heuristics, or using unsupervised reconstruction pre-training.  Furthermore, we demonstrated that
this phenomena is indeed an artifact of attention models -- that is, the suboptimal policies discovered by networks trained from scratch are suboptimal due to inefficiencies in their attention mechanisms.  This fact allows us to leverage pre-training in an  elegant way, by biasing a new model to copy the pre-trained model's glimpse patterns in early training stages.

Though a reasonable conclusion to draw from these results might be that pre-training happen to teach the model superior glimpse policies, we showed that this is not the case.  Rather, the high-efficiency glimpse policies were only obtained through a combination of pre-training (or bootstrapping) and task-specific training.  This led us to the hypothesis that any method that makes the model's attention mechanism more efficient than random chance in the early training batches has the potential to improve its ultimate performance, and in fact result in a more optimal glimpse policy than the one used for bootstrapping.  We tested this conjecture by implementing bootstrapped glimpse mimicking.  
We showed that constraining a randomly initialized network to copy a previously learned glimpse policy in early training stages can yield better performance.

A natural direction for future research is the extension of our methods to more complex classification tasks.  However, we are hopeful that insights derived from this paper may prove to be more generally applicable.  Though the methods of this work focused on attention models, one might reasonably conjecture that our conclusions could generalize to any model (for instance, one trained with reinforcement learning) that receives training inputs highly dependent on the model's own behavior. 

\subsubsection*{Acknowledgments}

Many thanks to Prof. Jay McClelland for supervising this research, Steven Hansen for providing helpful advice and technical assistance, Qihong Lu for discussing relevant ideas, and the Symbolic Systems program at Stanford for sponsoring the work.

\bibliographystyle{plainnat}
\bibliography{PreTrainingAttentionNIPS}

\small

\small

\end{document}